\documentclass[runningheads]{llncs}

\usepackage{eccv}

\usepackage{eccvabbrv}

\usepackage{graphicx}
\usepackage{booktabs}

\usepackage[accsupp]{axessibility}  

\usepackage{hyperref}

\usepackage{orcidlink}

\usepackage{multirow}
\usepackage{adjustbox}
\usepackage[table]{xcolor} 
\usepackage{xfrac}
\usepackage{array}
\usepackage{wrapfig}
\usepackage{colortbl}

\definecolor{downred}{RGB}{248,131,121}
\definecolor{upgreen}{RGB}{25,182,163}

\newcommand{\Real}{\mathbb{R}}

\renewcommand{\paragraph}[1]{\vspace{.4em}\noindent\textbf{#1}}

\newcommand{\ourfullname}{DVSM: Decoder-only View Synthesis Model Done Right\xspace}
\newcommand{\ourshortname}{DVSM\xspace}

\newcommand{\Image}{\mathbf{I}}
\newcommand{\Pose}{\mathbf{P}}
\newcommand{\Intr}{\mathbf{K}}

\newcommand{\rend}{\mathrm{Rend}}
\newcommand{\recon}{\mathrm{Recon}}
\newcommand{\plucker}{\mathrm{Pl\ddot{u}cker}}
\newcommand{\PErgb}{\mathrm{PE}^{\text{(rgb)}}}
\newcommand{\PEray}{\mathrm{PE}^{\text{(ray)}}}
\newcommand{\token}{\mathbf{X}}

\newcommand{\tokenrecon}{\token^{\text{(recon)}}}
\newcommand{\tokenrend}{\token^{\text{(rend)}}}

\begin{document}

\title{\ourfullname} 


\author{Cheng Sun$^{1}$ \quad
Jaesung Choe$^{1}$ \quad
Min-Hung Chen$^{1}$ \quad
Ryo Hachiuma$^{1}$ \quad
Yu-Chiang Frank Wang$^{1,2}$}

\authorrunning{C.~Sun et al.}

\institute{$^{1}$NVIDIA \quad $^{2}$National Taiwan University}

\maketitle

\begin{abstract}
Recent Large View Synthesis Models (LVSMs) advocate an encoder-decoder architecture that separates reconstruction and rendering into distinct networks. We re-examine this design. Through controlled experiments, we show that a decoder-only architecture, which represents scenes implicitly as a KV-cache, outperforms encoder-decoder variants while using fewer parameters at identical rendering complexity. Further analysis shows that sharing weights between the color-input reconstruction network and the camera-only rendering network better aligns their features at the same viewpoint, facilitating image synthesis. Building on this finding, our model, dubbed DVSM, further incorporates foundation model priors and stage-wise patch sizing for an improved efficiency-quality tradeoff. Our results establish a new state of the art for novel-view synthesis across multiple benchmarks, in some cases even outperforming per-scene-optimized 3DGS under dense input views.
\end{abstract}

\section{Introduction}
\label{sec:intro}



Novel view synthesis aims to render images from unobserved viewpoints given a set of input views. This task has traditionally relied heavily on explicit 3D inductive biases. For decades, the standard paradigm has involved reconstructing a 3D representation—such as meshes, neural radiance fields (NeRFs)~\cite{nerf}, or 3D Gaussian Splatting (3DGS)~\cite{3dgs}—and subsequently rendering novel views using a physics-inspired rendering equation (\eg, ray marching or splatting). In these conventional differentiable rendering pipelines, the scene optimization is deeply coupled with the rendering process; the reconstructed scene is fundamentally a function of the rendering function itself.
To be scalable with data, recent approaches like Large Reconstruction Model~\cite{lrm} employ neural networks to directly predict 3D representations from images, achieving promising results.

However, there are still some challenges exhibit in these 3D representation-based approaches.
For instance, the predicted 3D tends to fail reflective surface, mirror surface, complex material with volumetric scattering or subsurface scattering.
Although these are solvable by more advance renderer and 3D representations from graphic, incorporation with neural networks is still difficult.

Recently, the Large View Synthesis Model (LVSM)~\cite{lvsm} challenged this paradigm by proposing a fully data-driven, Transformer-based approach that minimizes 3D inductive biases. The scene representation and even the renderer are learned, which theoretically can adapt to any traditionally challenging surface or material given data. The LVSM framework explores two primary architectures: an encoder-decoder model that compresses input images into a fixed number of latent tokens (acting as a fully learned scene representation), and a decoder-only model that directly translates input multi-view tokens into target view tokens, completely bypassing any intermediate scene representation. While LVSM achieves state-of-the-art results by abandoning handcrafted 3D structures and renderer, we question whether all design choices are well-justified. Its encoder-decoder variant lacks the intrinsic link between reconstruction and rendering by using two separate networks with decoupled weights, where the encoder build scenes without directly using the knowledge in the decoder renderer. Its decoder-only variant builds scenes, represented as context tokens, independently for each novel-view query, which is counterintuitive and computationally costly.

We present Decoder-only View Synthesis Model Done Right (\ourshortname), demonstrating that a decoder-only architecture can be both efficient and performant. Rather than decoupling the encoder and decoder, or processing context views independently for each novel view without an intermediate representation, we represent scenes implicitly as a KV-cache. Crucially, we enforce strict weight sharing between the reconstruction stage, which builds the KV-cache, and the rendering stage, which queries it. Through systematic controlled experiments, we show that this design is essential: any decoupling of weights between the two stages leads to a drop in quality.

Interestingly, our design shares a similar philosophy with classical differentiable rendering methods. Approaches such as NeRF~\cite{nerf} and 3DGS~\cite{3dgs} employs differentiable renderer to reconstruce scenes. Our design inherits this spirit: our scenes as KV cache are also the outcome from a Transformer renderer, as reconstructor and renderer are the same network. In contrast, encoder-decoder models lack this property as their decoder renderer is merely a downstream module of a encoder reconstructor.

Building on this conceptually elegant and simple-yet-effective decoder-only design, we further advance the architecture by incorporating pre-trained foundation model priors and introducing a stage-wise patch sizing strategy to achieve superior efficiency-quality tradeoffs.
We summarize our contributions as follow:
\begin{itemize}
    \item \textbf{Efficient decoder-only architecture:}
    We propose a fully weight-shared, decoder-only framework that implicitly encodes 3D scenes as a KV-cache, achieving state-of-the-art quality with half the parameters of encoder-decoder designs at identical rendering complexity.
     \item \textbf{Insight on the superiority of decoder-only models:}
     Through thorough controlled experiments, we provide strong empirical evidence for this design. Feature-space analysis reveals better alignment between reconstruction and rendering stages, and an analogy to classical differentiable rendering further grounds our approach.
    \item \textbf{Foundation knowledge injection and stage-wise patch sizing:}
    We introduce two simple yet effective strategies--injecting foundation model knowledge and varying patch sizes across stages--that yield orthogonal improvements to the efficiency-quality trade-off.
\end{itemize}
\section{Related work}
\label{sec:related}

\paragraph{Scene optimization.}
Gradient-based optimization approaches have recently dominated the field of novel-view synthesis due to their outstanding quality.
Gradient-based optimization approaches have come to dominate novel-view synthesis thanks to their outstanding quality.
The differentiable scenes are optimized by reproducing the observed images via differentiable rendering.
Given enough observation, the per-scene trained scenes are shown to generalize to other viewpoints of the same scenes.
Neural Radiance Fields~\cite{nerf} is one earlier representative example, where scenes are defined as coordinate-based MLPs to serve volume rendering~\cite{Max}.
Subsequent works explore more efficient scene representations~\cite{dvgo,plenoxels,tensorf,instant-ngp,kilonerf}.
3DGS~\cite{3dgs} achieves next level of speed and quality by using Gaussian splats~\cite{ewa} as scenes and implementing CUDA-based rasterizer as their renderer.
Extensions have been made to ray tracing~\cite{3dgrt} as renderer and many other scene primitives~\cite{svraster,linprim,radiantfoam,meshsplat}.
However, these techniques perform much worse if views are not dense and the iterative per-scene optimization takes minutes to hours.
We study a generalizable approach, which works well even under few views and reconstruct fast by network feed-forwarding.

\paragraph{Geometric-based feed-forward models.}
Learning-based approaches are later explored so the model can keep improving with more training data.
Early approaches building cost volume to predict multi-plane images~\cite{mvsnet,pmsnet,gwcnet,cvpmvsnet,casmvsnet,llff} while limited to forward-facing viewing experience.
Methods predicting NeRF~\cite{mvsnerf,pixelnerf,ibrnet,neuray} cover a more general viewing angles.
After 3DGS~\cite{3dgs}, the problems are re-framed as pixel-aligned Gaussian attributes estimation task~\cite{pixelsplat,depthsplat,mvsplat,flash3d,lgm} for the neural networks, achieving faster and higher quality rendering.
Large Reconstruction Models (LRMs)~\cite{lrm} further advocates of using large Transformers~\cite{transformer} for its scalability.
LRM originally predict triplane~\cite{triplane} with following-up work extend to mesh~\cite{meshlrm} and GS~\cite{gslrm}.
However, these methods are limited to few input views.
Very recently, LongLRM~\cite{longlrm}, LongLRM++~\cite{longlrmpp}, and tttLRM~\cite{tttlrm} predict GS from dozen of input views by replacing the cross-view attention with Mamba2~\cite{mamba2} or Test-time Training~\cite{ttt,lact} layers.
Despite the promising results by geometric-based networks, there are still some fundamental limitation of the predicted geometry per se.
For instance, the GS fails to represent translucent material or mirror reflection surface.
Our geometric-free approach learn to render these challenging cases well. 
Our model is also applicable to longer sequence and achieve much better quality than the existing geometry-based methods.

\paragraph{Geometric-free feed-forward models.}
Another line of research explores neural-network as rendering function and side-steps the challenging 3D geometry reconstruction task.
For instance, some methods~\cite{attnrend,gpnr,lfnr} directly sample feature from reference views following epipolar geometry.
Some other methods~\cite{lfn,geofreevs,srt} define the rendered color as a neural network function of rays.
However, these models are limited by their network capacity.
Large View Synthesis Models (LVSMs)~\cite{lvsm} recently emerge with the powerful Transformer, achieving promising quality on several benchmarks.
Several works~\cite{efflvsm,lact,svsm} further improve LVSM architecture design, which we have a detail comparison after introducing our model.
Extensions of LVSMs cover unknown camera geometry setup~\cite{rayzer,erayzer,truesn,thely}, while we mainly focus on studying the fundamental aspect of LVSM under calibrated images.

\paragraph{Camera-aware generative models.}
Several modern generative models support controllable cameras~\cite{eschernet,syncdreamer,wonder3d,mvdiffusion,mvdiffusionpp,seva,viewcrafter,uni3c,recammaster,gen3c}.
Our study belong to the LVSM family with many goals not necessary covered by the generative ones: 1) we do not assume temporal order in rendering, 2) camera control are precise instead of discrete actions, 3) we aim to support dense context views for reconstruction, and 4) we assume target views can be reasoned from context. Despite that our regression-based model can not render unobserved region, we show better interpolate-viewpoints quality.
Incorporation with generative models is our future interest.

\section{Approach}
\label{sec:approach}

We give a task and framework overview in \cref{ssec:prelim}.
We detail our insight with a straightforward yet effective architecture, Decoder-only View Synthesis Model (DVSM), in \cref{ssec:arch}.
We further propose orthogonal improvements by leveraging the powerful pre-trained foundation model in \cref{ssec:prior_inject} and stage-wise patch sizing strategy to have a better computation-quality trade-off in \cref{ssec:patch_size}.

\subsection{Preliminary}
\label{ssec:prelim}

Our task is to synthesize novel views by inferring from the input calibrated images.
Specifically, the input is a set of $V$ images $\{\Image_i\in\Real^{H\times W\times 3}\}_{i=1}^V$ with their camera poses $\{\Pose_i\in \mathbb{SE}(3)\}_{i=1}^V$ and intrinsics $\{\Intr_i\in \Real^{3\times 3}\}_{i=1}^V$.
Our goal is to synthesize a novel view for random target camera queries: $(\Pose', \Intr')$.
Following the practice in this field, we do not assume any spatial ordering in the target queries and render each target view independently without using information from the other novel-view queries.
In this work, we also focus only on geometry inference, which assumes that target queries can mostly be synthesized by information from the given input views, instead of generation.

To this end, conventional methods reconstruct the underlying 3D either by per-scene optimization (\eg, NeRF~\cite{nerf}, 3DGS~\cite{3dgs}) or a generalizable network (recently known as LRM~\cite{lrm}). 
We study a different research line, Large View Synthesis Model (LVSM)~\cite{lvsm}, which recently emerges and renders target views by model feed-forwarding without explicitly predicting geometry representation.
Formally, the reconstruction and rendering processes become:
\begin{equation}
    \bar{\Image} = \rend(\Pose', \Intr', S ; \phi),  \quad 
    S = \recon(\Image, \Pose, \Intr ; \theta),
\end{equation}
where both $\recon(\cdot)$ and $\rend(\cdot)$ are neural networks parameterized by model weights $\theta$ and $\phi$, $\bar{\Image}$ is the synthesized novel-view, and $S$ is the predicted scenes implicitly represented as tokens~\cite{lvsm,svsm}, KV cache~\cite{efflvsm}, or updated fast weight~\cite{lact}.
During training, the model is trained to minimize the photometric deviation from the ground truth images $\Image'$ at the novel viewpoints:
\begin{equation}
    \mathcal{L} = \text{MSE}(\bar{\Image}, \Image') + \lambda \cdot \text{Percep}(\bar{\Image}, \Image')\,,
\end{equation}
where $\lambda$ is a weighting hyperparameters and $\text{Percep}$ is the perceptual loss~\cite{perceptual}.

\begin{figure}[tb]
\centering
\includegraphics[width=\linewidth]{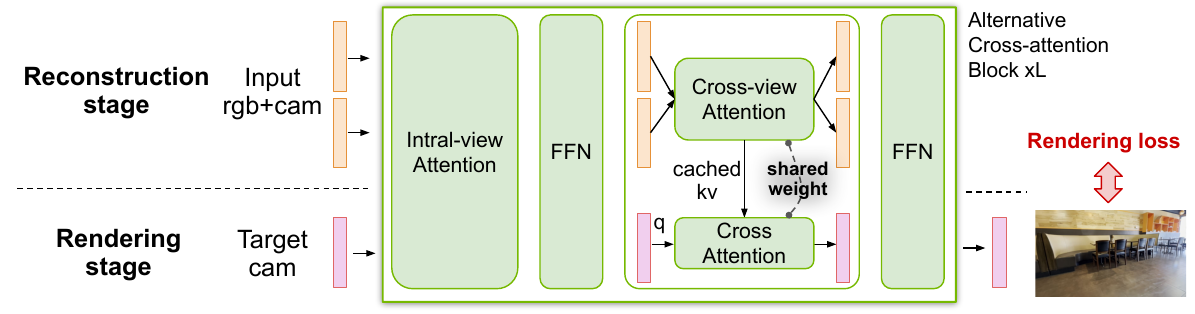}
\vspace{-2em}
\caption{{\bf Our architecture.}
Weights are fully shared between the reconstruction and rendering stages, including also the input tokenizer.
Please read \cref{ssec:arch} for our insight.
}
\label{fig:our_arch}
\vspace{-1em}
\end{figure}

\subsection{DVSM: an efficient decoder-only architecture}
\label{ssec:arch}

In contrast to the recent advocacy~\cite{efflvsm,svsm} of using an encoder and a decoder ViT~\cite{vit} as the $\recon(\cdot)$ and $\rend(\cdot)$ functions with two independent sets of model weights, our key finding is that a simple fully weight sharing between $\recon(\cdot)$ and $\rend(\cdot)$, including even the input patch embedding layers, is crucial for quality.
In the following, we describe our model first and then provide detail comparison with other architectures in the LVSM family later.

We start by input processing.
The tokens to the reconstruction stage mix color and camera information of patches, while the tokens to the novel-view rendering stage are composed of only the query camera information:
\begin{subequations}
\begin{align}
    \tokenrecon &= \text{LayerNorm}\left(\PEray(\plucker(\Pose, \Intr)) + \PErgb(\Image)\right)\,, \label{eq:xrecon}\\
    \tokenrend &= \text{LayerNorm}\left(\PEray(\plucker(\Pose', \Intr'))\right)\,,
\end{align}
\end{subequations}
where the $\plucker(\cdot)$ converts camera parameters to Plücker~\cite{plucker} ray maps with the same resolution as the images, $\PErgb(\cdot)$ and $\PEray(\cdot)$ are linear patch embedding with patch size $p$, and $\tokenrecon\in \Real^{V\times \frac{HW}{p^2} \times D}$ is the context tokens from the $V$ input views with latent dimension $D$, and $\tokenrend\in \Real^{\frac{HW}{p^2} \times D}$ is the novel-view tokens of a query target viewpoint. 
The weights of the two $\PEray(\cdot)$ and $\text{LayerNorm}(\cdot)$ are shared.

We illustrate our architecture in \cref{fig:our_arch}.
We use a single decoder ViT to process both $\tokenrecon$ and $\tokenrend$.
The decoder consists of a series of $L$ repeated blocks.
Inspired by VGGT~\cite{vggt}, we implement a block with an intra-view attention layer and a cross-view attention layer, both of which are followed by an MLP layer.
Residual connection~\cite{resnet} and QK-normalization~\cite{qknorm} are applied to all layers.
In the reconstruction stage, the keys and values of all the cross-view attention layers are cached~\cite{kvcache}, which can be viewed as an implicit scene representation to support novel-view queries.
In the rendering stage, $\tokenrend$ passes through the same decoder as for $\tokenrecon$.
The difference is only in the cross-view attention layer, where novel-view tokens use a query to retrieve scene information from the cached KV.
Finally, the novel-view tokens from the last layers $\tokenrecon_L$ are mapped to an image by:
\begin{equation}
    \bar{\Image} = \text{PixShuf}(\text{Linear}(\text{LayerNorm}(\tokenrend_L)))\,,
\end{equation}
where $\text{PixShuf}(\cdot)$ is pixel shuffling~\cite{pixshufl} that rearrange a latent to a image patch.

\begin{figure}[tb]
\centering
\includegraphics[width=\linewidth]{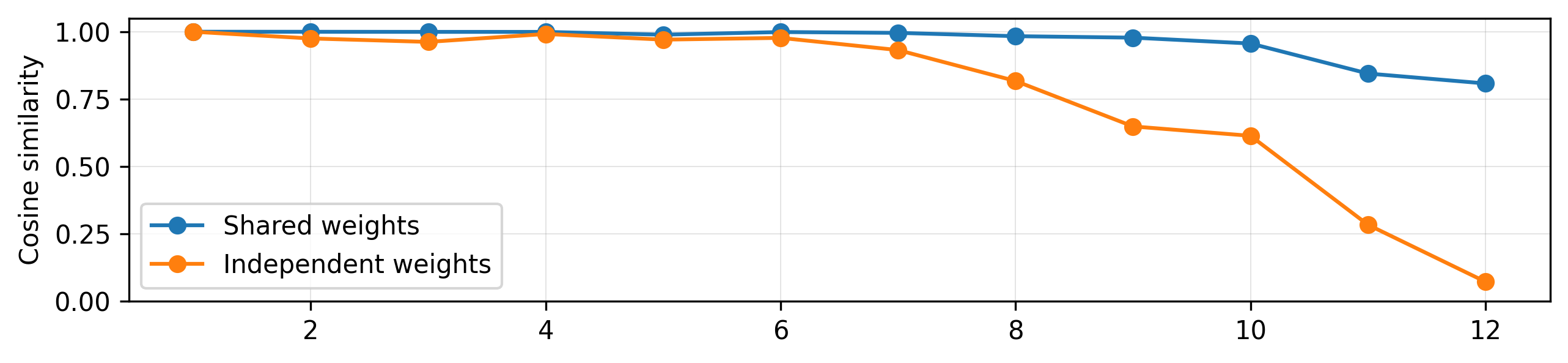}
\vspace{-2em}
\caption{{\bf Global features at cross-attention layers.}
We feed the same set of 32 views into the reconstruction (upper row) and the rendering branches (bottom row), and visualize results from the first frame.
We analyze the attended features, the $\text{softmax}(QK^T)V$, of cross attention layers, which is the only source for rendering branch to retrieve appearance information.
Both branches attend on the same set of $(K, V)$ and differ only in their queries $Q$.
Results show that the two branches without weight sharing retrieve more different global scene information under the same viewpoint.
}
\label{fig:attn_pca}
\vspace{-1em}
\end{figure}

\paragraph{Why decoder-only over encoder-decoder.}
Our detailed ablation studies show that decoupling any weights from the fully weight-shared decoder-only architecture leads to consistent drops in quality. Beyond these ablations, we offer two complementary perspectives.
First, from a representational standpoint, \cref{fig:attn_pca} compares the features retrieved at the same viewpoint during the reconstruction and rendering stages. Ideally, the rendering stage with only camera as input should retrieve scene features similar to those retrieved by the reconstruction stage, which has access to oracle color information. We find that features from the decoder-only model remain highly aligned across layers, whereas those from the encoder-decoder model diverge progressively in later layers.
Second, from a learning standpoint, the decoder-only model is effectively trained in a multi-task fashion: a single network receives gradients from both reconstruction and rendering objectives. In contrast, the encoder and decoder of an encoder-decoder model receive task-specific signals in isolation. No mechanism encourages them to at least converge to the shared weights as the decoder-only counterpart.


\begin{figure}[tb]
\centering
\includegraphics[width=\linewidth]{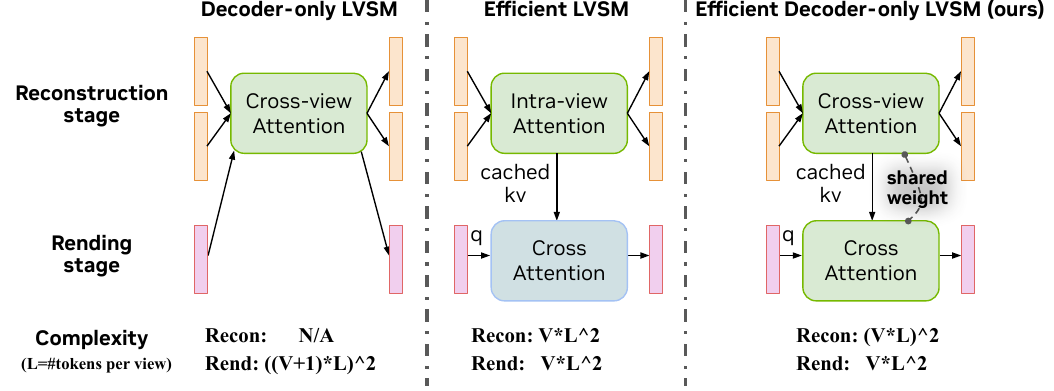}
\vspace{-2em}
\caption{{\bf Comparison of cross-attention layers.}
The last row denotes the complexity of a single cross-attention layer when processing the $V$ context input views in the reconstruction stage and processing a single camera viewpoint query in the rendering stage.
Decoder-only LVSM~\cite{lvsm} concatenates novel-view query tokens with all context tokens and feed-forward through the entire networks.
Efficient LVSM~\cite{efflvsm} processes each context views independently and represents scene as KV cache for the novel-view query to attend on.
We identify that model weight sharing is crucial and apply the same decoder in the two stages with KV caching to speedup.
We also use cross-view attention in the reconstruction stage for a better quality, which only increases reconstruction-stage time complexity while space complexity remain the same.
}
\label{fig:arch_comp}
\vspace{-1em}
\end{figure}

\paragraph{Comparison with LVSM-series.}
We illustrate the architectural difference with the most similar models in \cref{fig:arch_comp}.
\emph{Decoder-only LVSM}~\cite{lvsm} is an alternative architecture from the original LVSM paper, where the context views are duplicated and concatenated with every single novel-view query.
It not only substantially increases rendering time complexity but also implies that scenes (the context tokens) are reconstructed differently for each novel-view query (\cref{fig:arch_comp}'s the left-most), which is counterintuitive and inefficient.
Our decoder-only process context views once and caches their KV at cross-view attention layers for the rendering stage to retrieve.
\emph{BTimer}~\cite{btimer} works on dynamic scenes and uses decoder-only LVSM with KV-caching~\cite{kvcache} as a test-time data augmentator for explicit GS estimation, while it is only employed for a small-scale interpolation.
We focus on analyzing the fundamentals of LVSM on static and testing on large-scale scene reconstruction.
\emph{LaCT}~\cite{lact} use test-time training~\cite{ttt} to replace cross-view attention for efficiency.
\emph{Efficient LVSM}~\cite{efflvsm} caches KV of context views from an encoder with only intra-view self-attention, and uses another decoder to cross-attend on the cached KV (\cref{fig:arch_comp}'s the middle).
We find that employing weight-sharing in their encoder-decoder design is also helpful.
\emph{SVSM}~\cite{svsm} predicts scenes as dense-patch context tokens instead of the compressed global scene tokens from the original LVSM or the intermediate KV cache, as ours. The decoder then cross-attends on the context tokens.

\subsection{Foundation prior knowledge injection}
\label{ssec:prior_inject}

\begin{figure}[tb]
\centering
\includegraphics[width=\linewidth]{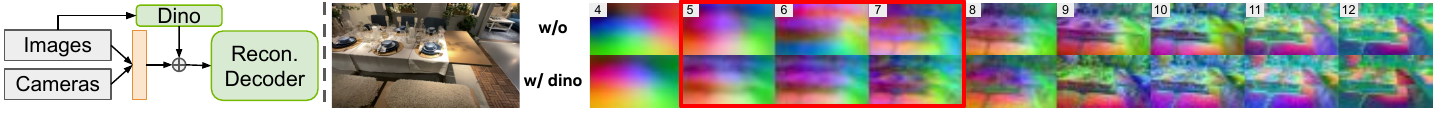}
\vspace{-2em}
\caption{{\bf Foundation feature injection.}
We propose a simple strategy to inject DINOv3~\cite{dinov3} feature into the reconstruction branch as shown in the left panel.
The rendering branch retrieves the foundation feature for novel-view queries from the cached KV and evolves into sharper feature on the earlier layers (the red highlight box).
}
\label{fig:dino_inject}
\vspace{-1em}
\end{figure}

The use of the powerful visual foundation models~\cite{dino,dinov2,dinov3,radio,radiov2} is not well explored in the field of LVSM.
We propose a simple and effective strategy to leverage such pretrained models into our efficient decoder-only architecture.
Specifically, we extend the patchifier \cref{eq:xrecon} in reconstruction branch into:
\begin{equation}
    \tokenrecon = \text{LayerNorm}\left(\PEray(\plucker(\Pose, \Intr)) + \PErgb(\Image) + \text{Prior}(\Image)\right)\,,
\end{equation}
where $\text{Prior}(\cdot)$ is the pretrained foundation models followed by a linear layer to project the number of latent dimensions.
We still keep the $\PErgb(\cdot)$ to emphasize more on the appearance information, as the foundation models are usually trained to be insensitive to color variation.
Such a simple strategy does not impact rendering efficiency.
The entire rendering computation cost remains the same, and the cached KVs are more enriched to retrieve, which helps sharpen the rendering tokens at earlier layers, as shown in \cref{fig:dino_inject}.

\subsection{Stage-wise patch-size}
\label{ssec:patch_size}

Conventionally, model depth and width are tweaked to offer different efficiency-quality tradeoffs.
We further explore another direction by using different patch sizes, $p_1\times p_1$ and $p_2\times p_2$, in the reconstruction and rendering stages, respectively.
For instance, applications typically can afford longer reconstruction time while the rendering FPS needs to be interactive.
We can train a model with $p_2{>}p_1$ to serve this purpose.
Our findings suggest that weight-sharing, even in the input patch embedding, is beneficial, so instead of implementing different channel projection layers for the two patch sizes, we resize the input to the receptive field of the target patch size.
Specifically, we set $q{=}\max(p_1, p_2)$ or to the patch size of the foundation model if employed, and resize 2D maps from $H{\times}W$ into $\lfloor\frac{H}{p_i}\rceil q {\times} \lfloor\frac{W}{p_i}\rceil q$.
Both the reconstruction and rendering stages use the same input projection layer with patch size $q$, and the resulting number of tokens is similar to that of applying $p_1^2$ and $p_2^2$ patchifier.

\begin{table}[tb]
  \caption{{\bf Ablation experiments.}
  All results are trained under a controlled environment with only modifications indicating by the ``Ablation'' column.
  We train with fewer batch size on 8 GPUs, and resolution curriculum is until a lower 720p resolution. 
  Model size report the number of trainable parameters.
  Efficiency is measured on a single A100.
  See \cref{ssec:abla} for details.
  {Markers: \colorbox{green!20}{better}, \colorbox{blue!10}{worse}, \colorbox{gray!20}{neutral} than (a).}
  }
  \label{tab:ablation}
  \vspace{-1em}
  \centering
  \begin{adjustbox}{max width=\linewidth}
  \begin{tabular}{@{}l@{\hskip 1pt}c@{\hskip 1pt}l c@{\hskip 10pt}ccc@{\hskip 10pt}cc@{}}
    \toprule
    \#
    && Ablation
    & Size(M)$\downarrow$
    & PSNR$\uparrow$ & SSIM$\uparrow$ & LPIPS$\downarrow$
    & Recon-time$\downarrow$ & Rend-fps$\uparrow$\\
    \midrule
    (a)&=& \ourshortname (ours)
    & 164 & 24.60 & 0.773 & 0.223 & 0.39s & 56.1 \\
    \midrule
    \multicolumn{9}{@{}l}{\bf \underline{Weight sharing}} \\
    (b)&=& (a)+decouple inp. proj.
    & \cellcolor{gray!10}166 & \cellcolor{blue!10}24.20 & \cellcolor{blue!10}0.757 & \cellcolor{blue!10}0.240 & \cellcolor{gray!10}0.39s & \cellcolor{gray!10}56.1 \\
    (c)&=& (a)+decouple intra-view attn.
    & \cellcolor{blue!10}191 & \cellcolor{blue!10}24.34 & \cellcolor{blue!10}0.763 & \cellcolor{blue!10}0.229 & \cellcolor{gray!10}0.39s & \cellcolor{gray!10}56.2 \\
    (d)&=& (a)+decouple cross-view attn. 
    & \cellcolor{blue!10}178 & \cellcolor{blue!10}24.06 & \cellcolor{blue!10}0.754 & \cellcolor{blue!10}0.240 & \cellcolor{gray!10}0.40s & \cellcolor{gray!10}55.7 \\
    (e)&=& (a)+decouple ffn
    & \cellcolor{blue!10}268 & \cellcolor{blue!10}23.89 & \cellcolor{blue!10}0.749 & \cellcolor{blue!10}0.240 & \cellcolor{gray!10}0.39s & \cellcolor{gray!10}56.3 \\
    (f)&=& (a)+decouple entire decoder
    & \cellcolor{blue!10}308 & \cellcolor{blue!10}23.19 & \cellcolor{blue!10}0.720 & \cellcolor{blue!10}0.268 & \cellcolor{gray!10}0.39s & \cellcolor{gray!10}56.1 \\
    \midrule
    \multicolumn{9}{@{}l}{\bf \underline{Cross-view attn. in reconstruction stage}} \\
    (g)&=& (a)+recon. no cross-view attn.
    & \cellcolor{gray!10}164 & \cellcolor{blue!10}22.98 & \cellcolor{blue!10}0.719 & \cellcolor{blue!10}0.267 & \cellcolor{gray!10}0.16s & \cellcolor{gray!10}56.9 \\
    (h)&=& (g)+decouple entire decoder
    & \cellcolor{blue!10}308 & \cellcolor{blue!10}21.89 & \cellcolor{blue!10}0.673 & \cellcolor{blue!10}0.305 & \cellcolor{gray!10}0.16s & \cellcolor{gray!10}57.3 \\
    \midrule
    \multicolumn{9}{@{}l}{\bf \underline{Foundation feature injection}} \\
    (i)&=& (a)+frozen DINOv3
    & \cellcolor{gray!10}164 & \cellcolor{green!20}25.12 & \cellcolor{green!20}0.789 & \cellcolor{green!20}0.206 & \cellcolor{blue!10}0.65s & \cellcolor{gray!10}55.7 \\
    (j)&=& (a)+tunable DINOv3
    & \cellcolor{blue!10}454 & \cellcolor{green!20}25.64 & \cellcolor{green!20}0.803 & \cellcolor{green!20}0.186 & \cellcolor{blue!10}0.65s & \cellcolor{gray!10}55.8 \\
    \midrule
    \multicolumn{9}{@{}l}{\bf \underline{Block composition}} \\
    (k)&=& (a)+no ffn in-between
    & \cellcolor{green!20}110 & \cellcolor{blue!10}24.05 & \cellcolor{blue!10}0.751 & \cellcolor{blue!10}0.246 & \cellcolor{gray!10}0.37s & \cellcolor{green!20}61.2 \\
    (l)&=& (a)+no intra-view attn.
    & \cellcolor{green!20} 83 & \cellcolor{blue!10}23.32 & \cellcolor{blue!10}0.725 & \cellcolor{blue!10}0.276 & \cellcolor{green!20}0.32s & \cellcolor{green!20}75.9 \\
  \bottomrule
  \end{tabular}
  \end{adjustbox}
\vspace{-1em}
\end{table}

\section{Experiments}
\label{sec:experiments}

\subsection{Controlled experiments}
\label{ssec:abla}

First, we show the controlled experiments that lead us to our final design choices.

\paragraph{Implementation details.} All experiments follow the same training scheduler and data sampler with 8 A100 GPUs, 32 context input views, 720p resolution, 768 latent dimensions, 12 cross-view attention layers, and a patch size of 16.
All variants are trained from scratch with a resolution curriculum of 340p, 480p, and finally to 720p for 100K, 10K, and 10K iterations, respectively.
We use the DL3DV~\cite{dl3dv} dataset for this experiment, which covers a wide variety of scenes.
On the 140 held-out testing scenes, we select every 8th frame as the target novel views and use K-Means to sample 32 out of the remaining frames (around 300 frames per scene on average) to serve as the input context views.

\paragraph{Weight sharing.}
Our model with fully-shared weights between the reconstruction and rendering stages is the base setup in \cref{tab:ablation}'s (a).
Results in \cref{tab:ablation}'s (b)--(f) show different levels of weight decoupling---meaning two different set of weights for (b) input patch embedding, (c) all intra-view attention layers, (d) the query and output projections in cross-view attention layers, (e) all fully-connected layers, and (f) the entire decoder, where the one for reconstruction stage is called encoder in this case.
The results clearly show the effectiveness of weight sharing, which uses fewer trainable parameters yet achieves the best quality.
Even decoupling the input projection layer can lead to an obvious drop in quality.
Using encoder-decoder (f) without any weight-sharing causes the most significant $-1.41$db PSNR degradation, despite having the largest number of trainable parameters.

\paragraph{Cross-view attention in reconstruction stage.}
We also test the proposed architecture by Efficient LVSM~\cite{efflvsm}, where all the cross-view attention layers perform intra-view attention instead in the reconstruction stage for efficiency.
The KVs of these layers are still cached for the corresponding layers in the rendering stage to cross-attend on.
The results are in \cref{tab:ablation}'s (g) with a weight-sharing decoder-only model and (h) with the original form of encoder-decoder design.
As expected, the reconstruction time is reduced by about $-60\%$, but the quality also significantly drops.
The rendering FPS is almost the same as the computation flow of (a), (g), and (h) are all identical in the rendering stage.
Although the reconstruction time is reduced, the quality degradation is non-negligible.
Thus we decide to keep the cross-view attention as it yields significant quality improvement.
Notably, weight sharing is still very helpful in such a model with $1.09$db PSNR difference.

\paragraph{Foundation feature injection.}
We inject features from DINOv3-L16~\cite{dinov3} into the reconstruction branch as in \cref{ssec:prior_inject}.
The improvements are $+0.52$db and $+1.04$db PSNR with and without finetuning the DINOv3, respectively.
The rendering computation flow remains the same with similar FPS.
The reconstruction time increases by $+66\%$ but is still quite affordable.

\paragraph{Block composition.}
We test the other block composition of (k) removing the fully-connected layers between intra- and cross-view attention, and (l) removing all the intra-view attention.
Despite having slightly better efficiency, since the quality drop is non-negligible, we keep the original design.

\begin{figure}[tb]
\centering
\includegraphics[width=.8\linewidth]{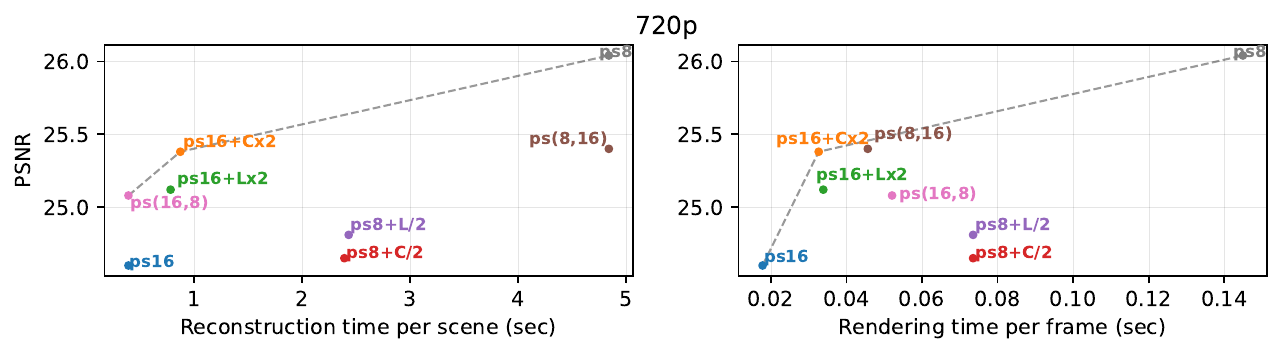}
\vspace{-1em}
\caption{{\bf Pareto front of model configurations.}
$ps$ with a single scalar is patch size; a tuple denotes the stage-wise patch sizes $(p_1, p_2)$ for reconstruction and rendering stages.
$C$ and $L$ are the number of latent channels and number of layers in our decoder, respective.
We explore models between the finest patch size of $8$ typically used in LVSM and LRM, and the efficient larger patch size of $16$ commonly used in other Transformers.
We show the reconstruction time per scene with 32 images and rendering time per frame.
Performance are evaluated under 720p resolution.
Pareto front are connected by the gray dash lines.
}
\label{fig:tradeoff_patch}
\vspace{-1em}
\end{figure}

\paragraph{Efficiency tradeoff.}
In contrast to the commonly used $14$ or $16$ patch size in most other Transformer models, we note that a finer patch size of $8$ is typically employed in the LVSM family.
Despite a good visual quality, it sacrifices rendering speed, which is relevant applications would be concern about.
We thus explore several model variations in \cref{fig:tradeoff_patch} in search of the Pareto front of an efficiency-quality tradeoff.
We adopt ps8 as the highest quality and ps16 as the most efficient setup, and tweak model width, depth, and the stage-wise patch size for several variants in-between.
Results are summarized as follows.
1) Increasing the number of channels for quality or decreasing the number of layers for speed achieves a better tradeoff.
2) Using finer rendering-stage patch size ($ps16{\rightarrow}ps(16,8)$ and $ps(8,16){\rightarrow}ps8$) significantly improve quality with identical reconstruction time.
3) Using a finer reconstruction-stage patch size ($ps(8,16)$) has a good rendering time tradeoff, especially if targeting lower resolution rendering.
Another merit of stage-wise patch-sizing is that it can save training time by adding them to the curriculum training schedule, as the parameter space of $ps16$, $ps(16,8)$, $ps(16,8)$, and $ps8$ are all identical.

\paragraph{Limitation.}
Several general Transformer advancements could bring further improvements, such as register tokens~\cite{regtoken}, projective rope~\cite{prope}, gating~\cite{gateattn}, and adaptive patch-size~\cite{apt}.
We skip exploring them due to time and resource limit.

\begin{table}[tb]
\begin{minipage}[t]{.46\linewidth}
  \caption{{\bf Two input views on Re10K.}
  }
  \label{tab:re10k}
  \centering
  \begin{adjustbox}{max width=\linewidth}
  \begin{tabular}{@{}l@{}r@{}c ccc@{}}
  \toprule
    Method & Venue && PSNR$\uparrow$ & SSIM$\uparrow$ & LPIPS$\downarrow$ \\
  \midrule
    \multicolumn{5}{@{}l}{\bf \underline{Resolution: 256$\times$256}} \\
    pixelNeRF~\cite{pixelnerf} & CVPR'21
    && 20.43 & 0.589 & 0.550 \\
    AttnRend~\cite{attnrend} & CVPR'23
    && 24.78 & 0.820 & 0.213 \\
    ViewCrafter~\cite{viewcrafter} & TPAMI'24
    && 21.63 & 0.642 & 0.175 \\
    GPNR~\cite{gpnr} & ECCV'24
    && 24.11 & 0.793 & 0.255 \\
    PixelSplat~\cite{pixelsplat} & CVPR'24
    && 26.09 & 0.863 & 0.136 \\
    MVSplat~\cite{mvsplat} & ECCV'24
    && 26.39 & 0.869 & 0.128 \\
    GS-LRM~\cite{gslrm} & ECCV'24
    && 28.10 & 0.892 & 0.114 \\
    DepthSplat~\cite{depthsplat} & CVPR'25
    && 27.46 & 0.889 & 0.115 \\
    SEVA~\cite{seva} & ICCV'25
    && 25.66 & 0.841 & 0.139 \\
    LVSM~\cite{lvsm} & ICLR'25
    && 28.58 & 0.893 & 0.114 \\
    Dec.-only~\cite{lvsm} & ICLR'25
    && 29.67 & 0.906 & 0.098 \\
    Btimer~\cite{btimer} & NeurIPS'25
    && 26.49 & 0.886 & \cellcolor{orange!25}0.096 \\
    Eff. LVSM~\cite{efflvsm} & ICLR'26
    && 28.93 & 0.895 & 0.102 \\
    SVSM~\cite{svsm} & arxiv'26
    && \cellcolor{orange!25}30.01 & 0.910 & \cellcolor{orange!25}0.096 \\
    \multicolumn{2}{@{}l}{\ourshortname (ours)\tiny{ ps16+dino}}
    && 29.68 & 0.904 & 0.101 \\
    \multicolumn{2}{@{}l}{\ourshortname (ours)\tiny{ ps8}}
    && 29.98 & \cellcolor{orange!25}0.912 & \cellcolor{orange!25}0.096 \\
    \multicolumn{2}{@{}l}{\ourshortname (ours)\tiny{ ps8+dino}}
    && \cellcolor{red!25}31.23 & \cellcolor{red!25}0.925 & \cellcolor{red!25}0.085 \\
  \midrule
    \multicolumn{5}{@{}l}{\bf \underline{Resolution: 512$\times$512}} \\
    LVSM~\cite{lvsm} & ICLR'25
    && 28.55 & 0.894 & 0.173 \\
    Dec.-only~\cite{lvsm} & ICLR'25
    && 29.53 & 0.904 & 0.141 \\
    Eff. LVSM~\cite{efflvsm} & ICLR'26
    && 29.86 & 0.905 & 0.147 \\
    \multicolumn{2}{@{}l}{\ourshortname (ours)\tiny{ ps16+dino}}
    && \cellcolor{orange!25}30.25 & 0.914 & \cellcolor{orange!25}0.137 \\
    \multicolumn{2}{@{}l}{\ourshortname (ours)\tiny{ ps8}}
    && 30.17 & \cellcolor{orange!25}0.915 & 0.138 \\
    \multicolumn{2}{@{}l}{\ourshortname (ours)\tiny{ ps8+dino}}
    && \cellcolor{red!25}31.17 & \cellcolor{red!25}0.924 & \cellcolor{red!25}0.129 \\
  \bottomrule
  \end{tabular}
  \end{adjustbox}
\end{minipage}%
\hfill
\begin{minipage}[t]{.52\linewidth}
  \caption{{\bf Many input views on DL3DV.}
  }
  \label{tab:dl3dv}
  \vspace{-1em}
  \centering
  \begin{adjustbox}{max width=\linewidth}
  \begin{tabular}{@{}l@{}r@{}c cc c ccc@{}}
    \toprule
    Method & Venue &&
    PSNR$\uparrow$ & SSIM$\uparrow$ & LPIPS$\downarrow$ &&
    Time$\downarrow$ & FPS$\uparrow$ \\
    \midrule
    \multicolumn{8}{@{}l}{\bf \underline{16 input views}}\vspace{.3em} \\
    3DGS~\cite{3dgs} & ToG'23 &&
    21.20 & 0.708 & 0.264 &&
    >600s & \cellcolor{red!25}>50 \\
    LVSM~\cite{lvsm}$^\dagger$ & ICLR'25 &&
    21.64 & 0.666 & 0.365 &&
    4s & 19.9 \\
    LongLRM~\cite{longlrm} & ICCV'25 &&
    22.66 & 0.740 & 0.292 &&
    \cellcolor{red!25}0.4s & \cellcolor{red!25}>50 \\
    LongLRM++~\cite{longlrmpp} & arxiv'25 &&
    24.40 & 0.795 & 0.231 &&
    1.6s & 14\\
    tttLRM~\cite{tttlrm} & CVPR'26 &&
    23.60 & 0.784 & 0.255 &&
    3.6s & -\\
    LaCT~\cite{lact} & ICLR'26 &&
    24.70 & 0.793 & 0.224 &&
    15s & 1.8\\
    \multicolumn{2}{@{}l}{\ourshortname (ours)\tiny{ ps16+dino}}
    &&
    \cellcolor{orange!25}25.46 & \cellcolor{orange!25}0.808 & \cellcolor{orange!25}0.199 &&
    \cellcolor{orange!25}0.6s & \cellcolor{orange!25}40 \\
    \multicolumn{2}{@{}l}{\ourshortname (ours)\tiny{ ps8}}
    &&
    \cellcolor{red!25}25.88 & \cellcolor{red!25}0.823 & \cellcolor{red!25}0.188 &&
    4.4s & 4 \\
    \midrule
    \multicolumn{7}{@{}l}{\bf \underline{32 input views}}\vspace{.3em} \\
    3DGS~\cite{3dgs} & ToG'23 &&
    23.60 & 0.779 & 0.213 &&
    >600s & \cellcolor{red!25}>50 \\
    LVSM~\cite{lvsm}$^\dagger$ & ICLR'25 &&
    21.73 & 0.664 & 0.365 &&
    15s & 20 \\
    LongLRM~\cite{longlrm} & ICCV'25 &&
    24.10 & 0.783 & 0.254 &&
    \cellcolor{red!25}1s & \cellcolor{red!25}>50 \\
    LongLRM++~\cite{longlrmpp} & arxiv'25 &&
    26.43 & 0.846 & 0.180 &&
    4.7s & 14\\
    tttLRM~\cite{tttlrm} & CVPR'26 &&
    25.07 & 0.822 & 0.215 &&
    7.2s & -\\
    LaCT~\cite{lact} & ICLR'26 &&
    26.90 & 0.837 & 0.185 &&
    29s & 1.8\\
    \multicolumn{2}{@{}l}{\ourshortname (ours)\tiny{ ps16+dino}}
    &&
    \cellcolor{orange!25}27.29 & \cellcolor{orange!25}0.852 & \cellcolor{orange!25}0.159 &&
    \cellcolor{orange!25}1.8s & \cellcolor{orange!25}25\\
    \multicolumn{2}{@{}l}{\ourshortname (ours)\tiny{ ps8}}
    &&
    \cellcolor{red!25}27.96 & \cellcolor{red!25}0.869 & \cellcolor{red!25}0.145 &&
    15s & 2\\
    \midrule
    \multicolumn{7}{@{}l}{\bf \underline{64 input views}}\vspace{.3em} \\
    3DGS~\cite{3dgs} & ToG'23 &&
    26.43 & 0.854 & 0.167 &&
    >600s & \cellcolor{red!25}>50\\
    LVSM~\cite{lvsm}$^\dagger$ & ICLR'25 &&
    21.46 & 0.651 & 0.377 &&
    60s & 19 \\
    LongLRM~\cite{longlrm} & ICCV'25 &&
    24.77 & 0.804 & 0.239 &&
    \cellcolor{red!25}3s & \cellcolor{red!25}>50\\
    LongLRM++~\cite{longlrmpp} & arxiv'25 &&
    27.30 & 0.869 & 0.161 &&
    16s & 14\\
    tttLRM~\cite{tttlrm} & CVPR'26 &&
    25.95 & 0.844 & 0.195 &&
    15s & -\\
    LaCT~\cite{lact} & ICLR'26 &&
    28.30 & 0.857 & 0.169 &&
    59s & 1.8\\
    \multicolumn{2}{@{}l}{\ourshortname (ours)\tiny{ ps16+dino}}
    &&
    \cellcolor{orange!25}28.90 & \cellcolor{orange!25}0.882 & \cellcolor{orange!25}0.135 &&
    \cellcolor{orange!25}5.2s & \cellcolor{orange!25}15\\
    \multicolumn{2}{@{}l}{\ourshortname (ours)\tiny{ ps8}}
    &&
    \cellcolor{red!25}29.71 & \cellcolor{red!25}0.898 & \cellcolor{red!25}0.120 &&
    60s & 1\\
    \midrule
    \multicolumn{7}{@{}l}{\color{gray}\bf \underline{Full input views (200--400 views)}}\vspace{.3em} \\
    \color{gray}3DGS~\cite{3dgs} & \color{gray}ToG'23 &&
    \color{gray}29.82 & \color{gray}0.919 & \color{gray}0.120 &&
    \color{gray}>600s & \color{gray}>50\\
  \bottomrule
  \end{tabular}
  \end{adjustbox}
\end{minipage} 
\vspace{-1em}
\end{table}

\subsection{Two input views evaluation}
\label{ssec:two_inp}

\paragraph{Benchmark.}
We use Re10K~\cite{stereomag} to evaluate our results on two input views.
The dataset contains 10K videos of real estate footage covering indoor and outdoor scenes.
Following LVSM~\cite{lvsm}, we use the train-test split from pixelSplat~\cite{pixelsplat}.
The baseline models include geometric-based~\cite{pixelnerf,pixelsplat,depthsplat,mvsplat,gslrm,btimer}, generative-based~\cite{viewcrafter,seva}, geometric-free~\cite{gpnr,attnrend,lvsm,efflvsm,svsm} approaches.

\paragraph{Implementation details.}
We use 768 latent channels with 12 Alternative Cross-attention Blocks (\cref{fig:our_arch}).
We train our models with patch sizes 8 and 16, both with DINOv3~\cite{dinov3} prior injection.
We train models on 256 resolution for 100K iterations with AdamW~\cite{adamw} and finetune on 512 resolution for 10K iterations.
Weight decay is set to $0.05$.
Learning rates are $4e{-}4$ and $1e{-}4$ with cosine annealing for the two resolutions.
Models are trained on 64 A100 GPUs.
In each iteration on a GPU, we sample 32 scenes with 2 context views and 3 target views.
Perceptual loss weight is set to $\lambda{=}0.2$.

Our implementation details mainly follow LVSM~\cite{lvsm}.
It is, however, costly to have results with all details aligning with the other competitive methods.
For instance, Efficient LVSM~\cite{efflvsm} use 1{,}024 latent channels and SVSM~\cite{svsm} trains with 170K iterations and advance camera encoding~\cite{prope}, while they only use one FFN in each block.
We suggest the reader take results from \cref{ssec:abla} with experiments under a fully controlled environment for our main claims.
The remaining comparisons are also affected by many other factors.

\paragraph{Results.}
We show a comparison in \cref{tab:re10k}.
Our model with $4\times$ fewer tokens (ps16) already outperforms most of the other methods, with comparable results to the original decoder-only LVSM~\cite{lvsm}.
When using the finest patch size (ps8) as the other methods do, our DINO-injected version achieves state-of-the-art quality and even outperforms the very recent Efficient LVSM~\cite{efflvsm} and SVSM~\cite{svsm}.
We report computational cost and qualitative results in the supplementary.

\begin{figure}[tb]
\centering
\includegraphics[width=\linewidth]{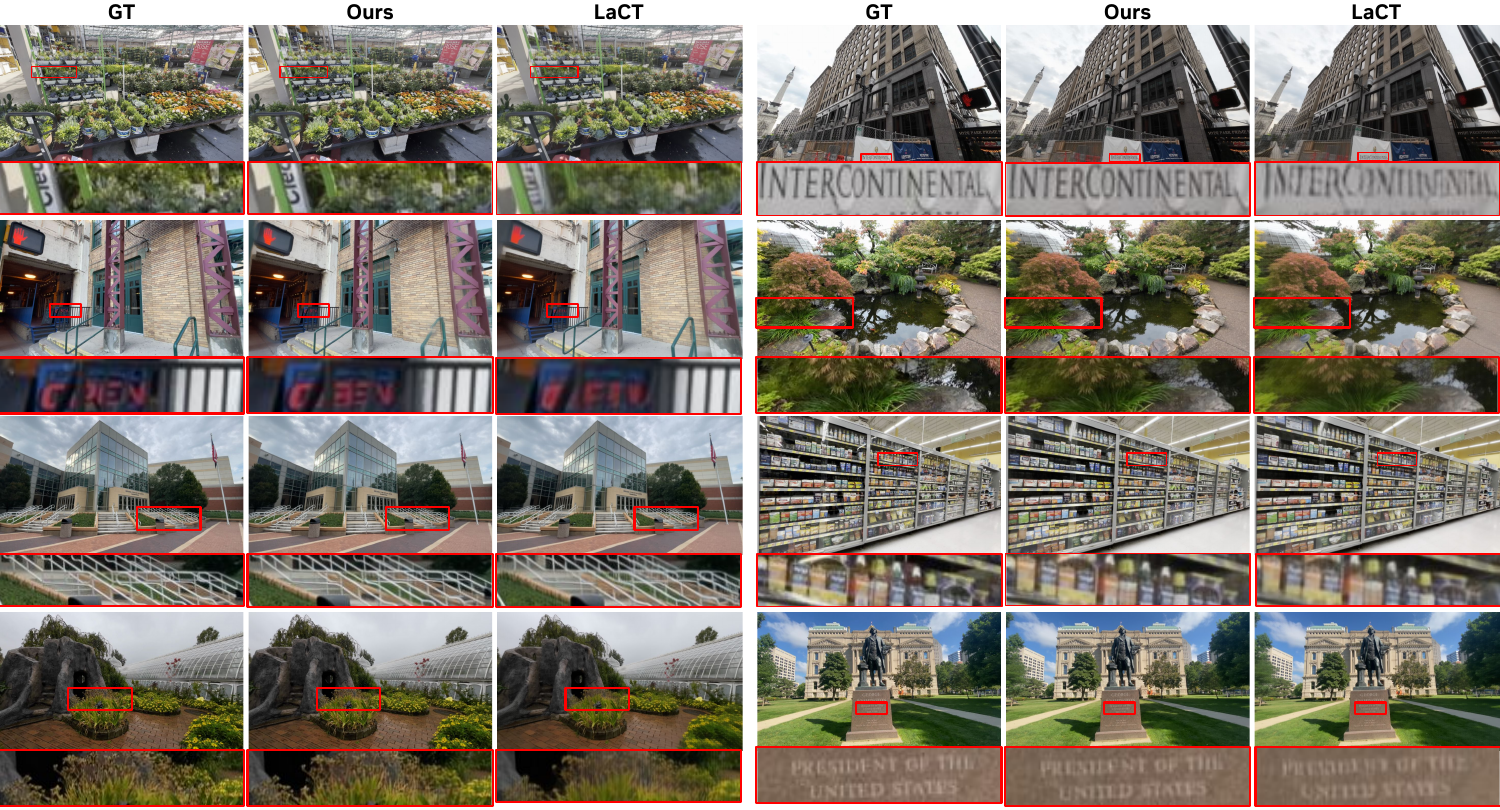}
\vspace{-1em}
\caption{{\bf Qualitative comparison.}
We show results from our model (ps8) and the most competitive baseline LaCT~\cite{lact}.
More results and videos are in the supplementary.
}
\label{fig:main_qual}
\vspace{-1em}
\end{figure}

\subsection{Dense input views evaluation}
\label{ssec:many_inp}

\paragraph{Benchmark.}
We use DL3DV~\cite{dl3dv} for denser input views evaluation.
There are around 10K scenes with diverse environments for training and 140 held-out scenes for testing.
We use the sampled testing view provided by LongLRM~\cite{longlrm}, which selects every 8-th frame in a scene as target views and uses K-means to sample 16, 32, and 64 context views from the non-target views.
The baseline methods include per-scene optimization-based~\cite{3dgs}, geometric-baesd~\cite{longlrm,longlrmpp,tttlrm}, and geometric-free~\cite{lvsm,lact} methods.
LVSM~\cite{lvsm} does not explore dense-view scenario so we re-train LVSM following our setup.

\paragraph{Implementation details.}
The implementation detail is similar to the one in \cref{ssec:two_inp}.
To perform better with many input views, we randomly sample 2, 4, 8, 16, and 32 context views from each scene during training.
The training curriculum is gradually increase resolution from 340p, 480p, 720p, to the final target 960p ($540{\times}960$).
The DINO~\cite{dino} prior is not employed for the ps8 version due to resource limitations.
More details are provided in the supplementary.

\paragraph{Results.}
We provide comparisons in \cref{tab:dl3dv}.
At such high-resolution, we find that our light-weight version with the larger patch size (ps16) already outperforms all previous methods.
For instance, our PSNR with ps16 is $+0.6$db more than the most competitive LaCT~\cite{lact} at 64 input views, while our reconstruction time is $10\times$ less and our rendering time is $8\times$ faster.
Our quality with ps8 significantly surpasses all previous methods. 
Notably, our method with only 64 input views already matches the quality of per-scene-optimized 3DGS using the full set of 200–400 input views.

\paragraph{Limitation.}
One common drawback of the LVSM family is the rendering speed compared to the other geometric-based approaches~\cite{longlrm}.
Even with ps16, our FPS is around $15$--$40$, which is not stable for a real-time application.
Our method also fails to process the full set of input views due to out of GPU memory issue.
Future work may want to explore general attention speedup~\cite{mqa,gqa}, reduce or merge the number of tokens~\cite{tome,apt,clift}, or other strategies to speed up rendering.

\paragraph{Qualitative results.}
We provide qualitative comparisons with LaCT~\cite{lact} in \cref{fig:main_qual}.
Our synthesized novel-view generally exhibits more high-frequency details from all scenes.
More visualizations are in the supplementary.

\begin{table}[t]
\caption{{\bf Zero-shot evaluations.} Metrics are PSNR$\uparrow$/SSIM$\uparrow$/LPIPS$\downarrow$.}
\label{tab:zeroshot}
\vspace{-1em}
\centering
\begin{adjustbox}{max width=\linewidth}
\begin{tabular}{@{}l @{\hskip 8pt} c @{\hskip 8pt} c @{\hskip 8pt} c@{}}
    \toprule
    Method
    & {\bf MipNerf360~\cite{mip-nerf360}}
    & {\bf Free~\cite{f2nerf}}
    & {\bf Hike~\cite{hike}} \\
    \midrule
    \multicolumn{4}{@{}l}{\bf \underline{16 input views}} \\
    LVSM {\scriptsize (173M)}
    & 19.69/0.454/0.557
    & 19.98/0.528/0.439
    & 18.48/0.453/0.494 \\
    LaCT {\scriptsize (379M)}
    & 20.10/0.494/0.498
    & 20.77/0.588/0.350
    & 18.35/0.512/0.403 \\
    Ours {\scriptsize\bf (164M)}
    & {\bf 21.76/0.564/0.413}
    & {\bf 22.04/0.664/0.285}
    & {\bf 20.24/0.591/0.363} \\
    \midrule
    \multicolumn{4}{@{}l}{\bf \underline{64 input views}} \\
    LVSM {\scriptsize (173M)}
    & 20.09/0.460/0.560
    & 20.27/0.517/0.444
    & 18.57/0.433/0.506 \\
    LaCT {\scriptsize (379M)}
    & 23.01/0.557/0.435
    & 24.61/0.680/0.265
    & 22.39/0.648/0.305 \\
    Ours {\scriptsize\bf (164M)}
    & {\bf 24.76/0.660/0.337}
    & {\bf 25.57/0.756/0.205}
    & {\bf 23.10/0.697/0.262} \\
    \bottomrule
\end{tabular}
\end{adjustbox}
\vspace{-1em}
\end{table}

\subsection{Generalization evaluation}
\label{ssec:generalization}

We further investigate model generalization ability from DL3DV~\cite{dl3dv} toward the other datasets: MipNerf360~\cite{mip-nerf360}, Free~\cite{f2nerf}, and Hike~\cite{hike} datasets.
We chose LVSM~\cite{lvsm} and the most competitive LaCT~\cite{lact} as our baselines.
Results in \cref{tab:zeroshot} show that our model (ps8) outperforms all baselines by a large margin on all datasets.
We find LVSM does not scale well with more input views, perhaps due to its fixed-size encoding tokens design.
Qualitative results in the supplementary are consistent with the numerical comparisons.

\paragraph{Limitation.}
All performances on unseen datasets are all in a worse scale compared to those in the in-domain evaluation \cref{tab:dl3dv}.
This suggests that the DL3DV dataset~\cite{dl3dv}, even with 10K scenes and 50M frames from diverse real-world scenes, alone is not enough for good generalizability.
Mix datasets training, as the other geometric-based foundation models~\cite{vggt,pi3,da3} would be helpful.

\begin{table}[tb]
\caption{{\bf Scannet++ iphone dataset~\cite{scannetpp}.} We submit our renderings to the official evaluation server. Our model infers from 128 views subsampled from the input videos. ``CC'' denotes color correction, applied to counter inconsistent lighting conditions.}
\label{tab:scannetpp_iphone}
\vspace{-1em}
\centering
\begin{tabular}{@{}l c ccc c ccc@{}}
    \toprule
    \noalign{\vspace{0.5ex}}
    Method &&
    PSNR$\uparrow$ & SSIM$\uparrow$ & LPIPS$\downarrow$ && 
    PSNR(CC)$\uparrow$ & SSIM(CC)$\uparrow$ & LPIPS(CC)$\downarrow$ \\
    \midrule
    3DGS~\cite{3dgs} &&
    15.170 & 0.823 & 0.425 && 20.027 & 0.850 & 0.349 \\
    Splatfacto~\cite{nerfstudio} && 
    14.113 & 0.810 & 0.435 && 19.356 & 0.850 & 0.360\\
    Ours &&
    {\bf 19.150} & {\bf 0.870} & {\bf 0.321} && {\bf 22.716} & {\bf 0.879} & {\bf 0.275}\\
    \bottomrule
\end{tabular}
\vspace{-1em}
\end{table}

\subsection{Extrapolated viewpoints under casual capturing}
\label{ssec:iphone}

\paragraph{Benchmark.}
We also evaluate on the ScanNet\texttt{++} iPhone dataset~\cite{scannetpp}. The videos are captured with default camera settings to reflect casual user capture. A held-out set of high-quality views is captured from extrapolated viewpoints rather than the commonly used interpolated viewpoints, providing a more challenging test setting.

\paragraph{Implementation details.}
We finetune our model, which is pretrained on DL3DV, on the training scenes of ScanNet\texttt{++}. At the time of submission, we are only able to train up to 480p resolution. For inference on held-out videos, we use 128 subsampled images, and the resulting renderings are directly upsampled to $1440\times 1920$ for evaluation.

\paragraph{Results.}
Our model significantly outperforms per-scene-optimized 3DGS, which uses the full video sequences. Without bells and whistles, our feed-forward model learns from data to reconstruct casually captured scenes despite challenges such as exposure variation, refocusing, and motion blur. In contrast, classical per-scene optimization requires hand-crafting a different strategy for each of these challenges.

\section{Conclusion}
\label{sec:final_remark}

We present \ourshortname, an efficient Decoder-only View Synthesis Model that performs novel-view synthesis in a geometry-free, feed-forward manner. Our controlled experiments and analysis demonstrate the importance of fully sharing weights between the reconstruction and rendering networks, which contradicts the recent advocacy of encoder-decoder designs in this field. \ourshortname achieves new state-of-the-art quality on several datasets, in some cases even matching or outperforming per-scene-optimized 3DGS trained on much more input views.

%
%
\bibliographystyle{splncs04}
\bibliography{main}

\vspace{1em}
\noindent\textbf{\Large Supplementary material}
\vspace{1em}

\setcounter{page}{1}
\setcounter{section}{0}
\renewcommand{\thesection}{\Alph{section}}
\renewcommand{\thetable}{\Alph{table}}
\renewcommand{\thefigure}{\Alph{figure}}
We provide additional implementation and dataset details in \cref{sec:supp_details}.
Efficiency report is summarized in \cref{sec:supp_runtime}.
Finally, more results are provided in \cref{sec:supp_qualitative}.

\section{Additional details} \label{sec:supp_details}

\subsection{Common implementation details}
We use the following setup for our final model training except stated otherwise.
Model latent dimension is set to 768 with 12 Alternative Cross-attention Blocks (main paper Fig.1).
We use AdamW optimizer with weight decay $0.05$ and momentum $(0.9, 0.95)$.
Learning rate scheduler follows a cosine curve with 2.5k linear warm-up steps.
Peak learning rate is $4e{-}4$ at the lowest resolution training and $1e{-}5$ for the higher-resolution finetuning.
Perceptual loss weight is set to $\lambda{=}0.2$.
All models are trained with 64 A100 GPUs.
Mixed-precision training with BFloat16 is activated when feed-forwarding the learnable layers.

We use the standard $\plucker$ function to embed camera information, which assigns a 6-dimensional $(\mathbf{r}_o{\times}\mathbf{r}_d, \mathbf{r}_d)$ vectors to each pixel, where $\mathbf{r}_o$ is the camera position, $\mathbf{r}_d$ is a unit vector indicating the pixel ray direction, and $\times$ is vector cross product operation.
The camera poses of the context views are normalized following GS-LRM~\cite{gslrm}'s strategy.

For all the transformer blocks, we use the pre-norm~\cite{prenorm} scheme with QK-normalization~\cite{qknorm}.
We use 12 attention heads and 4$\times$ MLP hidden dimension expansion.
We remove the bias term of all the linear projection layers.

\subsection{Two-view Re10K~\cite{stereomag} specific details}
In each iteration on a GPU, we sample 32 scenes each with 2 context views and 3 target views.
The frame skip between the 2 context views are randomly sampled in range of $[25, 192]$.
The target views are randomly sampled from the candidate frames consisting of all the intermediate views and extrapolation views with a maximum of $25$ frame skip of the two sampled context views.
We train models on $256{\times}256$ resolution for 100K iterations and $512{\times}512$ resolution for 10K iterations.

\subsection{Many views DL3DV~\cite{dl3dv} specific details}
To perform better with many input views, we randomly sample 2, 4, 8, 16, and 32 context views from each scene during training.
We sample the same number of target views as the context views for each scene.
The number of scenes of a training iteration is dynamically adjusted such that the total number of context views is 32 on a GPU.
The frame skip between the 2 nearby context views are randomly sampled in range of $[1, 16]$.
The target views are randomly sampled from the candidate set of $[\min(C)-2, \max(C)+2] - C$, where $C$ is the index set of the sampled context views and `$-$' is set subtraction.
A small subset of the scenes that are not following the 16:9 image aspect ratio and 960p image resolution are removed from training.
The lowest 340p resolution is trained with 100K iterations, while each of the higher 480p, 720p, and the final 960p ($540{\times}960$) resolutions are finetuned with 10K iterations.

\subsection{Zero-shot evaluation details}
We use our model trained on DL3DV with patch size 8 for the result in main paper's Table 4.
We use the 9 scenes from MipNerf360 dataset~\cite{mip-nerf360}, the 7 scenes from Free dataset~\cite{f2nerf}, and the 6 stable scenes suggested by \cite{longsplat} from Hike dataset~\cite{hike}.
Every 8-th frames of MipNerf360 and Free datasets and every 10-th frames of Hike dataset are selected as testing frames.
Context frames are sampled by K-means from the remaining frames.
We will release our view sampling split for future work to compare.
We resize all images to have around 518{,}400 (${=}540{\cdot}960$) pixels for model feed-forwarding.
The synthesized novel views are resized back to the source resolution for evaluation.
The same image processing procedure are applied to all methods.

\begin{table}[tb]
  \caption{{\bf Two input views on Re10K.}
  \underline{Recon-time} is the processing time of a single scene.
  \underline{Rend-fps} indicates the number of frames rendered per second without batching.
  \underline{Infer-mem} is the peak GPU memory usage when measuring the reconstruction and rendering time.
  \underline{Train-time/it} is the sum of model forward and backward time per iteration under training mode with a batch containing 32 context views for 16 scenes.
  \underline{Train-mem} is the peak memory usage when measuring training time.
  We \colorbox{green!10}{highlight} cells with numbers similar to the most efficient one under each column.
  }
  \label{tab:supp_re10k}
  \vspace{-1em}
  \centering
  \begin{adjustbox}{max width=\linewidth}
  \begin{tabular}{@{}l@{}r@{}c ccc |c cccccc@{}}
  \toprule
    Method & Venue && PSNR$\uparrow$ & SSIM$\uparrow$ & LPIPS$\downarrow$ &&
    \#params$\downarrow$ & Recon-time$\downarrow$ & Rend-fps$\uparrow$ & Infer-mem$\downarrow$ & Train-time/it$\downarrow$ & Train-mem$\downarrow$ \\
  \midrule
    \multicolumn{5}{@{}l}{\bf \underline{Resolution: 256$\times$256}} \\
    LVSM~\cite{lvsm} & ICLR'25
    && 28.58 & 0.893 & 0.114 &&
    \cellcolor{green!10}165M & \cellcolor{green!10}19ms & \cellcolor{green!10}73 & \cellcolor{green!10}1.0G & 2.4s & 16G \\
    Dec.-only LVSM~\cite{lvsm} & ICLR'25
    && 29.67 & 0.906 & 0.098 &&
    \cellcolor{green!10}165M & - & 22 & \cellcolor{green!10}1.2G & 5.1s & 25G \\
    Eff. LVSM~\cite{efflvsm} & ICLR'26
    && 28.93 & 0.895 & 0.102 &&
    199M & \cellcolor{green!10}17ms & 68 & \cellcolor{green!10}1.2G & 0.9s & \cellcolor{green!10}9G \\
    \multicolumn{2}{@{}l}{\ourshortname (ours)\tiny{ ps16+dino}}
    && 29.68 & 0.904 & 0.101 &&
    454M & 45ms & 57 & 2.7G & \cellcolor{green!10}0.5s & \cellcolor{green!10}9G\\
    \multicolumn{2}{@{}l}{\ourshortname (ours)\tiny{ ps8}}
    && \cellcolor{orange!25}29.98 & \cellcolor{orange!25}0.912 & \cellcolor{orange!25}0.096 &&
    \cellcolor{green!10}163M & \cellcolor{green!10}20ms & 59 & \cellcolor{green!10}1.1G & 1.1s & \cellcolor{green!10}10G\\
    \multicolumn{2}{@{}l}{\ourshortname (ours)\tiny{ ps8+dino}}
    && \cellcolor{red!25}31.23 & \cellcolor{red!25}0.925 & \cellcolor{red!25}0.085 &&
    454M & 44ms & 58 & 2.9G & 2.0s & 17G \\
  \midrule
    \multicolumn{5}{@{}l}{\bf \underline{Resolution: 512$\times$512}} \\
    LVSM~\cite{lvsm} & ICLR'25
    && 28.55 & 0.894 & 0.173 &&
    \cellcolor{green!10}165M & 50ms & 36 & \cellcolor{green!10}1.2G & 6.3s & 28G\\
    Dec.-only LVSM~\cite{lvsm} & ICLR'25
    && 29.53 & 0.904 & 0.141 &&
    \cellcolor{green!10}165M & - & 6 & 1.5G & 21s & 60G\\
    Eff. LVSM~\cite{efflvsm} & ICLR'26
    && 29.86 & 0.905 & 0.147 &&
    199M & 50ms & 41 & 1.6G & 5.8s & 27G\\
    \multicolumn{2}{@{}l}{\ourshortname (ours)\tiny{ ps16+dino}}
    && \cellcolor{orange!25}30.25 & 0.914 & \cellcolor{orange!25}0.137 &&
    454M & \cellcolor{green!10}45ms & \cellcolor{green!10}58 & 2.9G & \cellcolor{green!10}2.0s & \cellcolor{green!10}17G\\
    \multicolumn{2}{@{}l}{\ourshortname (ours)\tiny{ ps8}}
    && 30.17 & \cellcolor{orange!25}0.915 & 0.138 &&
    \cellcolor{green!10}163M & 57ms & 36 & 1.6G & 6.5s & 34G\\
    \multicolumn{2}{@{}l}{\ourshortname (ours)\tiny{ ps8+dino}}
    && \cellcolor{red!25}31.17 & \cellcolor{red!25}0.924 & \cellcolor{red!25}0.129 &&
    454M & 136ms & 36 & 3.4G & 11s & 51G\\
  \bottomrule
  \end{tabular}
  \end{adjustbox}
\end{table}

\begin{table}[tb]
  \caption{{\bf Many input views on DL3DV with $544{\times}960$ resolution.}
  \underline{Recon-time} is the processing time of a single scene.
  \underline{Rend-fps} indicates the number of frames rendered per second without batching.
  \underline{Infer-mem} is the peak GPU memory usage when measuring the reconstruction and rendering.
  \underline{Train-time/it} is the sum of model forward and backward time per iteration under training mode with a batch containing 32 context views.
  \underline{Train-mem} is the peak memory usage when measuring training time.
  We do not report training measurement under 64 input views as we sample a maximum of 32 input views per scene during training.
  Besides, the actual training randomly sample different number of input views per scene instead of a fixed number.
  We \colorbox{green!10}{highlight} cells with numbers similar to the most efficient one under each column.
  }
  \label{tab:supp_dl3dv}
  \vspace{-1em}
  \centering
  \begin{adjustbox}{max width=\linewidth}
  \begin{tabular}{@{}l@{}r@{}c ccc|c cccccc@{}}
    \toprule
    Method & Venue &&
    PSNR$\uparrow$ & SSIM$\uparrow$ & LPIPS$\downarrow$ &&
    \#params$\downarrow$ & Recon-time$\downarrow$ & Rend-fps$\uparrow$ & Infer-mem$\downarrow$ & Train-time/it$\downarrow$ & Train-mem$\downarrow$ \\
    \midrule
    \multicolumn{8}{@{}l}{\bf \underline{16 input views}}\vspace{.3em} \\
    LVSM~\cite{lvsm} & ICLR'25 &&
    21.64 & 0.666 & 0.365 &&
    \cellcolor{green!10}165M & 3.8s & 20 & \cellcolor{green!10}5G & 41s & 42G \\
    LaCT~\cite{lact} & ICLR'26 &&
    24.70 & 0.793 & 0.224 &&
    379M & 2.2s & 10 & 13G & 34s & 74G \\
    \multicolumn{2}{@{}l}{\ourshortname (ours)\tiny{ ps16+dino}}
    &&
    \cellcolor{orange!25}25.46 & \cellcolor{orange!25}0.808 & \cellcolor{orange!25}0.199 &&
    454M & \cellcolor{green!10}0.6s & \cellcolor{green!10}40 & \cellcolor{green!10}5G & \cellcolor{green!10}8s & \cellcolor{green!10}28G \\
    \multicolumn{2}{@{}l}{\ourshortname (ours)\tiny{ ps8}}
    &&
    \cellcolor{red!25}25.88 & \cellcolor{red!25}0.823 & \cellcolor{red!25}0.188 &&
    \cellcolor{green!10}163M & 4.4s & 4 & 11G & 73s & 65G \\
    \midrule
    \multicolumn{7}{@{}l}{\bf \underline{32 input views}}\vspace{.3em} \\
    LVSM~\cite{lvsm} & ICLR'25 &&
    21.73 & 0.664 & 0.365 &&
    \cellcolor{green!10}165M & 15s & 20 & \cellcolor{green!10}9G & 73s & 42G\\
    LaCT~\cite{lact} & ICLR'26 &&
    26.90 & 0.837 & 0.185 &&
    379M & 4.3s & 10 & 23G & 36s & 72G \\
    \multicolumn{2}{@{}l}{\ourshortname (ours)\tiny{ ps16+dino}}
    &&
    \cellcolor{orange!25}27.29 & \cellcolor{orange!25}0.852 & \cellcolor{orange!25}0.159 &&
    454M & \cellcolor{green!10}1.8s & \cellcolor{green!10}25 & \cellcolor{green!10}8G & \cellcolor{green!10}12s & \cellcolor{green!10}28G \\
    \multicolumn{2}{@{}l}{\ourshortname (ours)\tiny{ ps8}}
    &&
    \cellcolor{red!25}27.96 & \cellcolor{red!25}0.869 & \cellcolor{red!25}0.145 &&
    \cellcolor{green!10}163M & 15s & 2 & 20G & 133s & 65G\\
    \midrule
    \multicolumn{7}{@{}l}{\bf \underline{64 input views}}\vspace{.3em} \\
    LVSM~\cite{lvsm} & ICLR'25 &&
    21.46 & 0.651 & 0.377 &&
    \cellcolor{green!10}165M & 60s & \cellcolor{green!10}20 & 17G & - & -\\
    LaCT~\cite{lact} & ICLR'26 &&
    28.30 & 0.857 & 0.169 &&
    379M & 8.7s & 10 & 44G & - & -\\
    \multicolumn{2}{@{}l}{\ourshortname (ours)\tiny{ ps16+dino}}
    &&
    \cellcolor{orange!25}28.90 & \cellcolor{orange!25}0.882 & \cellcolor{orange!25}0.135 &&
    454M & \cellcolor{green!10}5.2s & 15 & \cellcolor{green!10}14G & - & -\\
    \multicolumn{2}{@{}l}{\ourshortname (ours)\tiny{ ps8}}
    &&
    \cellcolor{red!25}29.71 & \cellcolor{red!25}0.898 & \cellcolor{red!25}0.120 &&
    \cellcolor{green!10}163M & 60s & 1 & 40G & - & -\\
  \bottomrule
  \end{tabular}
  \end{adjustbox}
\end{table}

\section{Runtime report} \label{sec:supp_runtime}

We report more efficiency related metrics comparing to LVSM-based methods in \cref{tab:supp_re10k,tab:supp_dl3dv}.
We use a unified protocol to measure the time and space usage of all methods.
Specifically, we create random input to feed into models and take the median measurement of 5 runs.
We disable \emph{torch.compile} for both inference and training measurements.
Activation checkpointing are applied to each transformer block for all methods when measuring training efficiency.
Efficiency results of LaCT are evaluated with our re-implementation so the numbers are different from the main paper, which are adopt from their official report.

\paragraph{Efficiency impact by injecting DINO feature.}
Employing DINO largely increase the number of trainable parameters and training/inference resource usage, except rendering FPS.
As discussed in the main paper, the DINO is only used in reconstruction stage and the rendering computational flow remains identical to the one without using DINO.

\paragraph{Model size.}
Our model variant without the DINO has similar number of parameters comparing to the baseline LVSM, Decoder-only LVSM, and Efficient LVSM, while using much less parameters than LaCT.
However, the quality of it outperforms all the baseline by a clear margin.
Note that a larger model does not imply lower efficiency as it also highly related to the number of tokens (derived from patch sizes) and the model time complexity for processing tokens.
Our model with the larger patch size 16 and DINO has more parameters than the other baseline methods while it is the most efficient one under several metrics and setups.

\begin{table}
  \centering
  \caption{{\bf COLMAP-free setup.} We show a simple extension to use feed-forward camera poses instead of the offline computed camera parameters by COLMAP. The other methods in this table estimate camera poses instead of using COLMAP poses.}
  \label{tab:nopose}
  \vspace{-1em}
  \centering
  \begin{tabular}{@{}l@{}r@{}c ccc@{}}
  \toprule
    Method & Venue && PSNR$\uparrow$ & SSIM$\uparrow$ & LPIPS$\downarrow$ \\
  \midrule
    Splatt3R~\cite{splatt3r} & arxiv'24
    && 15.32 & 0.490 & 0.436 \\
    CoPoNeRF~\cite{coponerf} & CVPR'24
    && 18.94 & 0.619 & 0.388 \\
    NoPoSplat~\cite{noposplat} & ICLR'25
    && \cellcolor{orange!25}25.03 & \cellcolor{orange!25}0.838 & \cellcolor{orange!25}0.160 \\
    \multicolumn{2}{@{}l}{\ourshortname (ours) {\tiny +$\pi^3$~\cite{pi3} poses}}
    && \cellcolor{red!25}27.55 & \cellcolor{red!25}0.860 & \cellcolor{red!25}0.124 \\
  \bottomrule
  \end{tabular}
\end{table}

\section{Additional results} \label{sec:supp_qualitative}

\paragraph{COLMAP-free extension.}
We show a straightforward extension to use a feed-forward camera estimator instead of COLMAP.
To compare with previous COLMAP-free methods on Re10k dataset, we use our variant with patch size 8 and DINO prior variant.
We use $\pi^3$~\cite{pi3} as a off-the-shelf tool to estimate camera poses.
Following the setting of the baseline NoPoSplat~\cite{noposplat}, we use the ground-truth camera intrinsic.
The testing scene split is the same as the one in main paper's \cref{tab:re10k} but the sampling views follow the ones from NoPoSplat.

The comparison is presented in \cref{tab:nopose}.
We show that our simple combination outperforms previous dedicated models under this setup.
Future work may consider building on top of our pipeline for a COLMAP-free and geometric-free feedforward novel-view synthesis model.

\begin{figure}[tb]
\centering
\includegraphics[width=\linewidth]{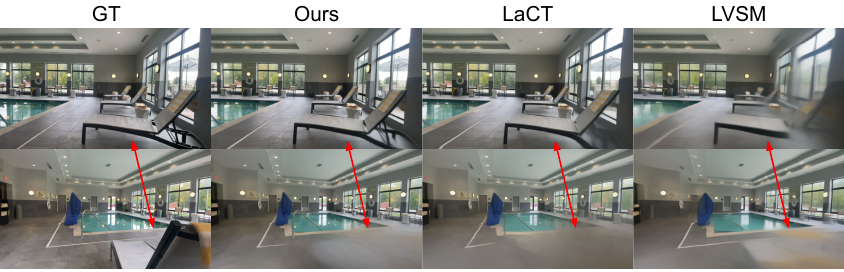}
\vspace{-1em}
\caption{{\bf Failure case.}
}
\label{fig:supp_failure}
\end{figure}

\begin{figure}[tb]
\centering
\includegraphics[width=\linewidth]{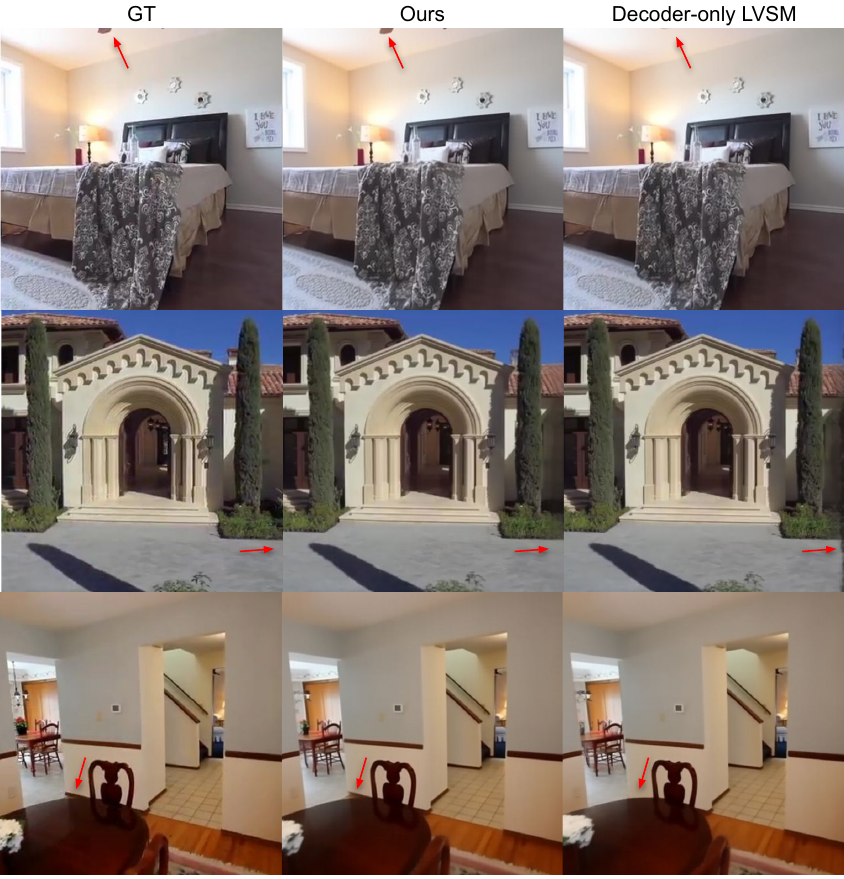}
\vspace{-1em}
\caption{{\bf Qualitative results on Re10K dataset~\cite{stereomag} with two context views.}
}
\label{fig:supp_re10k}
\end{figure}

\begin{figure}
\centering
\includegraphics[width=\linewidth]{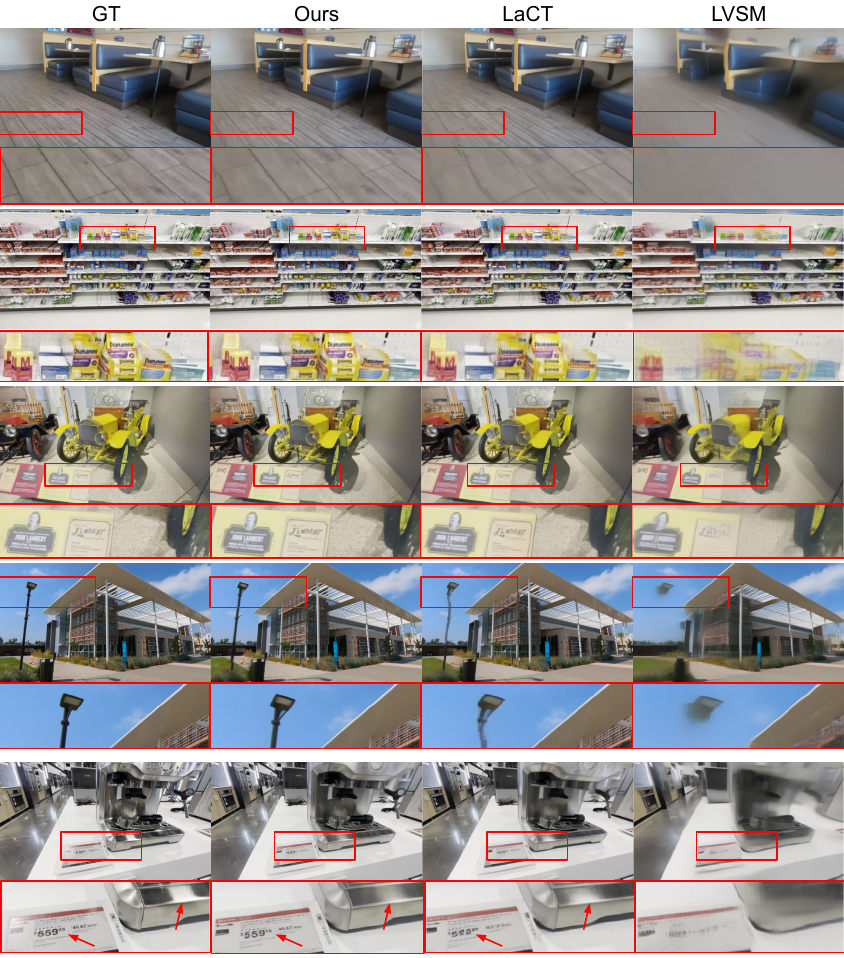}
\vspace{-1em}
\caption{{\bf Qualitative results on DL3DV dataset~\cite{dl3dv} with 64 context views.}
}
\label{fig:supp_dl3dv}
\end{figure}

\begin{figure}
\centering
\includegraphics[width=.9\linewidth]{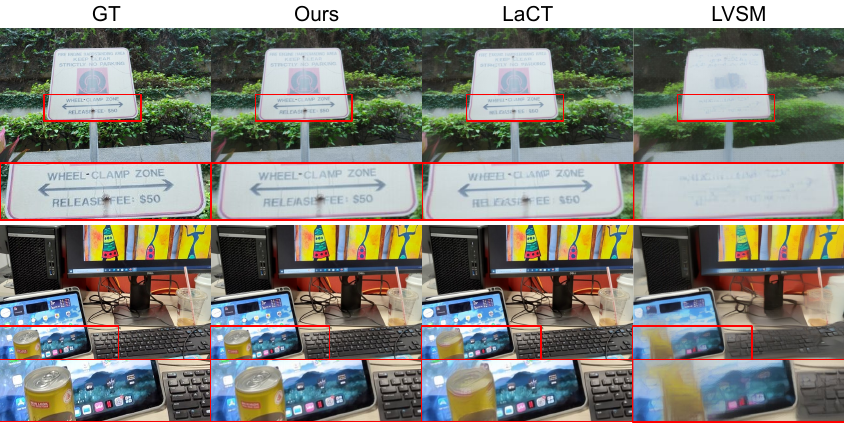}
\vspace{-1em}
\caption{{\bf Zero-shot results on Free dataset~\cite{f2nerf} with 64 context views.}
}
\label{fig:supp_free}
\end{figure}

\begin{figure}
\centering
\includegraphics[width=.9\linewidth]{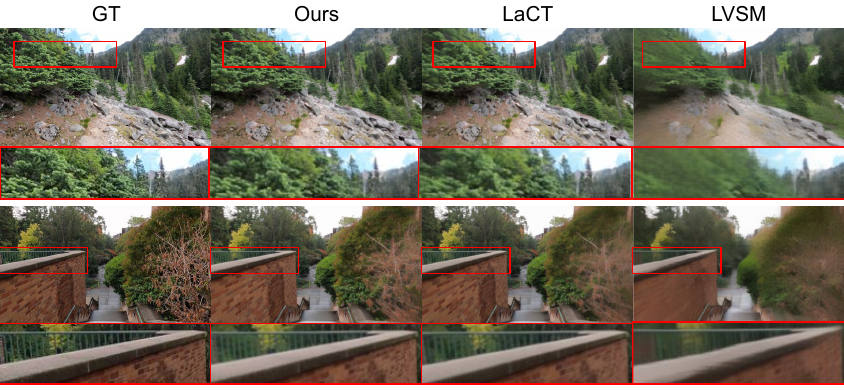}
\vspace{-1em}
\caption{{\bf Zero-shot results on Hike dataset~\cite{hike} with 64 context views.}
}
\label{fig:supp_hike}
\end{figure}

\begin{figure}[tb]
\centering
\includegraphics[width=.9\linewidth]{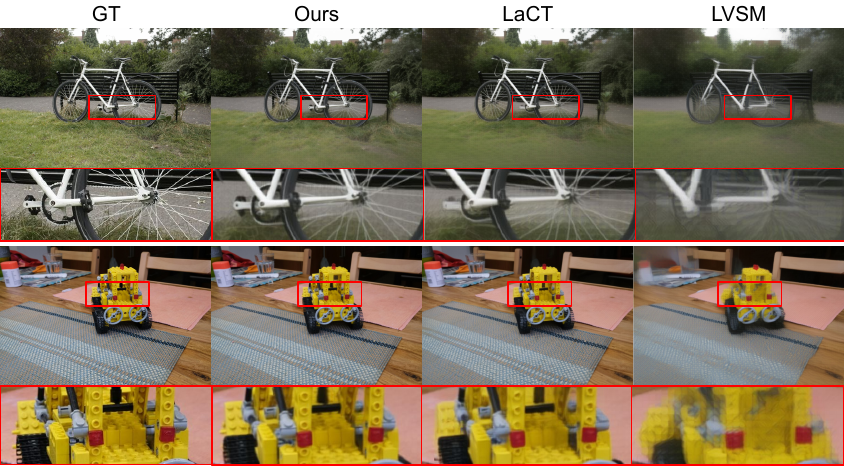}
\vspace{-1em}
\caption{{\bf Zero-shot results on MipNerf360 dataset~\cite{mip-nerf360} with 64 context views.}
}
\label{fig:supp_mipnerf360}
\end{figure}

\paragraph{Qualitative results.}
We provide qualitative results in \cref{fig:supp_failure,fig:supp_re10k,fig:supp_dl3dv,fig:supp_free,fig:supp_hike,fig:supp_mipnerf360}. Please also see the attachment for video results.

In \cref{fig:supp_failure}, we highlight a failure case of our method when the camera poses extrapolate too much from the context views.
In this example, despite the lounge chair is rendered correctly under the viewpoint of the first row, it suddenly disappears from another viewpoint in the second row.
We observe that it is also a common failure mode of LVSM series of methods, where the incorrectly reconstructed geometry is synthesized into an unpredictable appearance.
Future work may want to design advanced view sampling strategy to improve geometric and view extrapolation.

Qualitative comparisons in \cref{fig:supp_re10k,fig:supp_dl3dv,fig:supp_free,fig:supp_hike,fig:supp_mipnerf360} show that our model synthesize sharper views with geometrically correct details (\eg, the text part).
The results reflect the quantitative results in the main paper.

\end{document}